\documentclass[conference]{IEEEtran}
\IEEEoverridecommandlockouts

\usepackage{threeparttable}
\usepackage{amsmath}
\usepackage{amssymb}
\usepackage{mathtools}
\usepackage{amsthm}
\usepackage{booktabs}
\usepackage{amsmath}
\usepackage{amsthm}
\usepackage{amsfonts}
\usepackage{amssymb}
\usepackage{graphicx}
\usepackage{mathrsfs}
\usepackage{float}
\usepackage{tikz}
\usepackage{tikz-network}
\usepackage{pgfplots}
\usepackage{colortbl}
\usepackage{enumitem}
\usepackage{xspace}
\usepackage{algorithm}
\usepackage[noend]{algpseudocode}
\usepackage{caption}
\setlist[description]{leftmargin=\parindent,labelindent=\parindent}

\newcommand{\swag}{SWAG\xspace}
\newcommand{\lga}{LGA\xspace}
\usetikzlibrary{decorations.pathreplacing,calc,shapes,positioning,arrows,arrows}
\usepackage{pgfplots}
\usepackage{subcaption}
\pgfplotsset{%
	,compat=1.12
	,every axis x label/.style={at={(current axis.right of origin)},anchor=north west}
	,every axis y label/.style={at={(current axis.above origin)},anchor=north east}
}
\usetikzlibrary{datavisualization,shapes.geometric,arrows.meta,decorations.markings, fit}
\usetikzlibrary{datavisualization.formats.functions}
\definecolor{scarlet}{rgb}{1.0, 0.13, 0.0}
\definecolor{brightmaroon}{rgb}{0.76, 0.13, 0.28}
\definecolor{mediumturquoise}{rgb}{0.28, 0.82, 0.8}
\definecolor{fandango}{rgb}{0.71, 0.2, 0.54}
\definecolor{antiquewhite}{rgb}{0.98, 0.92, 0.84}
\definecolor{babyblue}{rgb}{0.54, 0.81, 0.94}
\definecolor{brilliantlavender}{rgb}{0.96, 0.73, 1.0}
\definecolor{bronze}{rgb}{0.8, 0.5, 0.2}
\definecolor{cornsilk}{rgb}{1.0, 0.97, 0.86}
\definecolor{lavenderpink}{rgb}{0.98, 0.68, 0.82}
\definecolor{sandybrown}{rgb}{0.96, 0.64, 0.38}
\definecolor{celadon}{rgb}{0.67, 0.88, 0.69}

\newcommand{\sg}{\text{sg}}
\newcommand{\relu}{\text{ReLU}}
\newcommand{\ber}{\text{Ber}}

\newcommand{\unif}{\text{Unif}}
\newcommand{\iid}{\overset{\text{i.i.d}}{\sim}}

\newcommand{\result}[2]{${#1}_{\pm#2}$}
\newcommand{\resultf}[2]{${\color{red}\mathbf{#1}_{\pm#2}}$}
\newcommand{\results}[2]{${\color{blue}\underline{#1}_{\pm#2}}$}

\DeclareMathOperator*{\concat}{%
    \mathchoice%
        {\Big\Vert}%
        {\big\Vert}%
        {\Vert}%
        {\Vert}%
}

\theoremstyle{plain}

\theoremstyle{definition}

\theoremstyle{remark}

\theoremstyle{plain}

\makeatletter
\newcommand{\linebreakand}{%
  \end{@IEEEauthorhalign}
  \hfill\mbox{}\par
  \mbox{}\hfill\begin{@IEEEauthorhalign}
}
\makeatother
\begin{document}
    \title{Self-supervision meets kernel graph neural models:\\ From architecture to augmentations}
    \author{
        \IEEEauthorblockN{Jiawang Dan\IEEEauthorrefmark{2}\IEEEauthorrefmark{1}, Ruofan Wu\IEEEauthorrefmark{2}\IEEEauthorrefmark{1}\thanks{\IEEEauthorrefmark{1} Equal contribution}, Yunpeng Liu\IEEEauthorrefmark{2}\IEEEauthorrefmark{3}, Baokun Wang\IEEEauthorrefmark{2}}
        \IEEEauthorblockN{Changhua Meng\IEEEauthorrefmark{2}, Tengfei Liu\IEEEauthorrefmark{2}, Tianyi Zhang\IEEEauthorrefmark{2}, Ningtao Wang\IEEEauthorrefmark{2}, Xing Fu\IEEEauthorrefmark{2}, Qi Li\IEEEauthorrefmark{3}, Weiqiang Wang\IEEEauthorrefmark{2}}
        \IEEEauthorblockA{\IEEEauthorrefmark{2}Ant Group}
        \IEEEauthorblockA{\IEEEauthorrefmark{3}Tsinghua University}
        \IEEEauthorblockA{\{yancong.djw, ruofan.wrf, lyp402072, yike.wbk, aaron.ltf, ningtao.nt\}@antgroup.com}
        \IEEEauthorblockA{qli01@tsinghua.edu.cn, \{changhua.mch, zty113091, zicai.fx, weiqiang.wwq\}@antgroup.com}
    }

    \maketitle
    \begin{abstract}
        Graph representation learning has now become the de facto standard when handling graph-structured data, with the framework of message-passing graph neural networks (MPNN) being the most prevailing algorithmic tool. Despite its popularity, the family of MPNNs suffers from several drawbacks such as transparency and expressivity. Recently, the idea of designing neural models on graphs using the theory of graph kernels has emerged as a more transparent as well as sometimes more expressive alternative to MPNNs known as kernel graph neural networks (KGNNs). Developments on KGNNs are currently a nascent field of research, leaving several challenges from algorithmic design and adaptation to other learning paradigms such as self-supervised learning. In this paper, we improve the design and learning of KGNNs. Firstly, we extend the algorithmic formulation of KGNNs by allowing a more flexible graph-level similarity definition that encompasses former proposals like random walk graph kernel, as well as providing a smoother optimization objective that alleviates the need of introducing combinatorial learning procedures. Secondly, we enhance KGNNs through the lens of self-supervision via developing a novel structure-preserving graph data augmentation method called \emph{latent graph augmentation (LGA)}. Finally, we perform extensive empirical evaluations to demonstrate the efficacy of our proposed mechanisms. Experimental results over benchmark datasets suggest that our proposed model achieves competitive performance that is comparable to or sometimes outperforming state-of-the-art graph representation learning frameworks with or without self-supervision on graph classification tasks. Comparisons against other previously established graph data augmentation methods verify that the proposed LGA augmentation scheme captures better semantics of graph-level invariance. 
    \end{abstract}

    \section{Introduction}\label{sec: intro}
    Recent years have witnessed surging developments in graph representation learning (GRL) which utilizes neural models over graph-structured data for downstream tasks like node classification \cite{kipf2016semi} and graph classification \cite{xu2018powerful}. Among these the task of graph classification has shown promising results in applications like drug discovery \cite{stokes2020deep} and protein function prediction \cite{gligorijevic2021structure}. The design of GRL algorithms has attracted significant interest, with the idea of message-passing graph neural networks \cite{pmlr-v70-gilmer17a} (hereafter abbreviated as MPNNs) being the de facto choice. In its most standard form, MPNNs obtain node representations via aggregating neighborhood information for each node in a recursive manner, with an optional combination mechanism at each step. The node representations are further aggregated into a graph-level representation for downstream tasks like graph classification \cite{xu2018powerful}. In the seminal work \cite{xu2018powerful}, the authors explored the expressivity limits of message-passing protocols and concluded that the expressivity limit of such kinds of procedures is bounded by first-order Weisfeler-Leman isomorphism tests. While there exist more theoretically expressive variants such as higher-order graph neural networks \cite{morris2021weisfeiler}, these alternatives are typically computationally expensive and lose the scalability of MPNNs. \par
    In a recent line of works \cite{nikolentzos2020random, feng2022kergnns, nikolentzos2023geometric}, the authors have studied an alternative way of defining neural models over graphs which are inspired by the theory of graph kernels \cite{vishwanathan2010graph}, we will call such variants kernel graph neural networks (KGNNs). KGNNs use a set of trainable hidden graphs and use the similarity of the input graph with the hidden graphs as the building block of graph representations. KGNNs provide a powerful alternative to MPNNs in the following aspects: Firstly, KGNNs allow more transparent graph modeling, since the obtained hidden graphs themselves are easily visualized, they provide interpretations of the learned model. Secondly, KGNNs share several important properties with MPNNs such as permutation invariance and scalability. Finally, it was proved in previous works \cite{feng2022kergnns} that KGNNs are provably more expressive than MPNNs. Despite these advantages, direct optimization over hidden graphs is computationally intractable. In the original proposal \cite{nikolentzos2020random}, the authors represent hidden graphs as continuous-valued matrices, resulting in subtle definitions. It is thus of interest to further explore this modeling paradigm via reinspecting the similarity definition, and its relation to the theory of graph kernels.\par
    Another recent trend in the area of GRL is the development of self-supervision techniques \cite{wu2021self}, as they are particularly effective in label-scarce scenarios, providing a principled way of facilitating unlabelled data. In self-supervision, one typically constructs a pretext task whose supervisory signals are derived through a certain sense of \emph{invariance}, and utilizes some invariance-promoting objectives to obtain pre-trained models such as contrastive objectives \cite{you2020graph} and non-contrastive objectives \cite{thakoor2021bootstrapped}. While there has been abundant work on self-supervision for MPNNs \cite{wu2021self}, so far as we have noticed there have been no studies on self-supervision over KGNNs. Applying self-supervision to the (pre)training of KGNN is challenging since KGNNs do not produce \emph{intermediate node-level representations} which are necessary for many self-supervised GRL frameworks \cite{sun2019infograph, suresh2021adversarial, hou2022graphmae}. To perform (invariance-based) self-supervised learning with KGNNs, we need to construct our supervision signals via graph-level invariant transforms instead of node-level operations. Although common practices in graph data augmentations \cite{you2020graph} also apply to KGNNs, they might not be able to express the right semantics of invariance that could be captured by KGNN architectures: As KGNNs are intuitively understood to focus more on graph structures, we hypothesize that successful pretraining of KGNN networks require \emph{structure-preserving} graph data augmentations, which has yet been under-explored in the current literature.\par
    In this paper, we make several contributions to the design and learning of KGNNs which are elaborated on as follows:
    \begin{itemize}[leftmargin=*]
        \item Our first contribution is a novel extension of the current KGNN design that allows a more flexible definition than contemporary KGNNs which are mostly derived through random walk graph kernels. We also instantiate a concrete model that utilizes the technique of graph diffusion \cite{gasteiger2019diffusion}. The proposed model is intuitively understood as a smoothed version of contemporary KGNNs.
        \item Our second contribution is a novel structure-preserving graph data augmentation method that serves as a building block for self-supervised pre-training using KGNNs. The semantics of structure-preserving is expressed through a wide range of random graph models, as well as being adaptive to individual instances.
        \item Finally, we conduct extensive empirical evaluations under the task of graph classification over benchmark datasets. Experimental results suggest our proposed model improves over contemporary KGNNs, and sometimes achieves performance on par with or even outperforms state-of-the-art methods. 
    \end{itemize}
    \section{Methodology}
    \begin{figure*}
        \centering
        \includegraphics[width=0.7\textwidth]{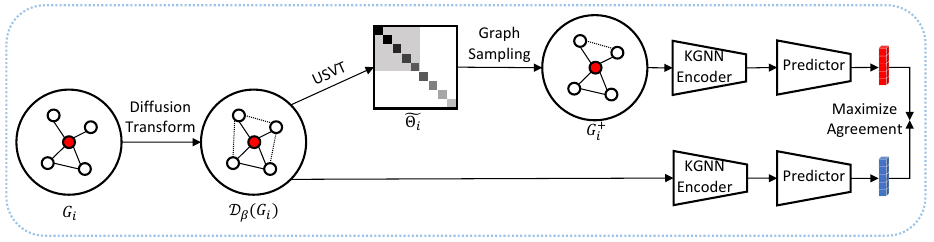}
        \caption{A concise illustration of the proposed self-supervised graph representation learning framework that consists of two main components: The first one is the \swag architecture that utilizes graph diffusion transform to allow continuous optimization. The second one is the \lga method for graph data augmentation that reflects the notion of structural invariance which is beneficial for the learning of KGNNs.}
        \label{fig: model_architecture}
    \end{figure*}
    \subsection{Kernel graph neural networks (KGNNs)}
    Let $G = (V, E) $ be an undirected graph associated with node features $X \in \mathbb{R}^d$. We will use $V_G$ and $E_G$ to denote the node set and edge set of graph $G$, further denote $[N]$ as the set $\{1, \ldots, N\}$. Given a training set $\mathbf{G} = \{G_1, \ldots, G_N\}$ of $N$ graphs that are of varying sizes with each one having a label $Y_i, i \in [N]$, we are interested in learning a graph representation extractor $h$ that maps graph objects to euclidean vectors, as well as learning a task-specific downstream model $g$ (sometimes referred to as \emph{predictor}) that maps graph representations to predictions. In supervised learning paradigms, $h$ and $g$ are trained in an end-to-end fashion, with $h$ usually being certain kinds of graph neural networks and $g$ being an MLP. In self-supervised learning paradigms, $h$ and $g$ are trained separately, with $h$ being trained using supervision signals that are not directly associated with the downstream task label. After obtaining the pre-trained encoder, we either train $g$ using probing (i.e., freezing the pre-trained graph representations) or using fine-tuning techniques. In this paper, we will focus on self-supervised approaches with $f$ being parameterized using KGNNs. Specifically, a KGNN model typically involves a collection of $M$ \emph{learnable hidden graphs} as trainable parameters, i.e., $\mathcal{H} = \{H_m\}_{m \in [M]}$. Graph representations are obtained via computing some similarity metric between the input graph and all the hidden graphs, with the prevailing practice being inspired by the theory of graph kernels \cite{vishwanathan2010graph}. 
    With the most representative architecture derived from the theory of \emph{random walk kernels} \cite{nikolentzos2020random, feng2022kergnns}. 
    Given input graphs $G$ and $G^\prime$, the random walk kernel evaluation $K(G, G^\prime)$ computes a weighted sum of the counts of random walks of varying lengths that the two graphs have in common. The kernel definition allows incorporating node features \cite{nikolentzos2020random} via treating pairwise feature similarity as additional weights. Formally, given a walk length $p \ge 1$, the $p$-th constituent of random walk kernel between $G$ and $G^\prime$ is computed as
    \begin{align}\label{eqn: rw_kernel}
        K^{(p)}(G, G^\prime) = \sum_{k, l \in V_{G^\prime}} \sum_{i, j \in V_G} \langle x_i, x^\prime_k \rangle A^p_{ij} {A^\prime}^p_{kl} \langle x_j, x^\prime_l \rangle,
    \end{align}
    where we define $A$ and $A^\prime$ as the adjacency matrix of $G$ and $G^\prime$, and $x^\prime_u, u \in V_{G^\prime}$ as the node feature of graph $G^\prime$. To evaluate graph kernels, we may compute a weighted sum (ideally infinite) over all $ K^{(p)}(G, G^\prime), p > 1$. In practice, it was observed that computing $p$ up to a small to moderate integer (i.e., $p \le 3$) suffices for most applications, and the representations are constructed via concatenating all $K^{(p)}(G, G^\prime)$ with $ 1\le p \le P$ and $G^\prime \in \mathcal{H}$. 
    \subsection{Extensions and smoothed random walks}
    According to the original definition \eqref{eqn: rw_kernel}, the trainable parameters involve a collection of (hidden) feature matrices and a collection of (hidden) adjacency matrices. However, adjacency matrices are binary-valued and thus do not fit into the standard gradient-based learning framework in neural models. Besides, directly optimizing potentially large binary matrices will incur a combinatorial level of complexity and is computationally intractable. To fix this issue, in \cite{nikolentzos2020random} the authors suggested replacing $A^\prime$ in \eqref{eqn: rw_kernel} by $\relu(W)$ with $W$ being an arbitrary continuous-valued matrix. While this fix addresses the computational issue, it breaks the semantics of graph kernels (as the resulting formula does not count the common random walks). However, since exact kernel evaluations may not contribute significantly to the performance in downstream tasks, we might directly extend the definition of \eqref{eqn: rw_kernel} so that it allows smoother optimization formulations, in particular, we propose the following extension
    \begin{align}\label{eqn: smooth_kernel}
        \resizebox{0.91\linewidth}{!}{$
        \widetilde{K}^{(p)}(G, G^\prime) = \sum_{k, l \in V_{G^\prime}} \sum_{i, j \in V_G} s_K (x_i, x^\prime_k ) B^p_{ij} {B^\prime}^p_{kl} s_K( x_j, x^\prime_l ),
        $}
    \end{align}
    where $s_K$ denotes some similarity function that applies to euclidean vectors,  $B$ and $B^\prime$ are some \emph{continuous-valued characteristics} of graphs $G$ and $G^\prime$, respectively. The above extension naturally serves as a concrete similarity measure, while at the same time allowing direct gradient-based optimization. In this paper, we will stick to the sleekest similarity formulation of $s_K$ which is the inner product function. What remains is to find a proper characteristic $B$, which in this paper we choose to be the diffusion matrix of graph $G$ \cite{gasteiger2019diffusion}.
    \begin{align}\label{eqn: diffusion}
        \mathcal{D}_\beta(G) = \sum_{j=0}^\infty \beta_j T^j,
    \end{align}
    where $\beta_j$ stands for some coefficient configurations that make the summation \eqref{eqn: diffusion} convergent and $T$ is a transition matrix defined as $D^{-1/2} A D^{-1/2}$, with $D$ being the diagonal matrix with diagonal values corresponding to node degrees. For $B^\prime$ in \eqref{eqn: smooth_kernel}, we may simply parameterize it as a continuous-valued symmetric matrix with values in $[0, 1]$. We term the resulting encoder as \textbf{S}moothed random \textbf{WA}lk \textbf{G}raph neural network (SWAG), whose encoding function with maximum walk length $P > 1$ and the number of hidden graphs $M$ is given by:
    \begin{align}
        h_\text{\swag} (G) = \concat_{m \in [M], p \in [P]} \widetilde{K}^{(p)}(G, H_m),
    \end{align}
    where we use $\concat$ to denote the concatenation operation. We use $d_\text{o} = MP$ to denote the dimension of output graph representation. 
    
    \subsection{Self-supervision via struture-preserving graph data augmentation}\label{sec: ssl}
    In this paper, we are mainly interested in self-supervised learning primitives using \emph{augmentations}, with two canonical frameworks being contrastive learning and non-contrastive learning. Both frameworks are based on an augmentation mechanism. Under the context of GRL, for any input graph $G$ (sometimes referred to as \emph{anchor graph}), we augment it with a positive sample $G^+ \sim \mathcal{A}(\cdot | G)$, where we use $\mathcal{A}(\cdot | G)$ to denote the (conditional) augmentation distribution given the anchor graph $G$ that is supported on the universe of graphs $\mathcal{G}$. \par
    \textbf{Constrastive pre-training} In contrastive frameworks, the learning objective is derived via contrasting augmented views (positive samples) with a set of negative samples $ \{G^-_1, \ldots, G^-_K \}$, each being sampled from the (conditional) distribution $\mathcal{N}(\cdot | G)$. We will adopt the InfoNCE objective \cite{oord2018representation}:
    \begin{align}\label{eqn: infonce}
        \resizebox{0.91\linewidth}{!}{$
        \mathcal{L}_{\text{cst}} = \frac{1}{N} \sum_{G \in \mathbf{G}} \dfrac{e^{s_L( h(G), h(G^+))}}{e^{s_L( h(G), h(G^+))} + \sum_{k \in [K]} e^{s_L( h(G), h(G^-_k))}},
        $}
    \end{align}
    with $s_L$ being some similarity function. We will be using the form $s_L(x, y) = \langle \phi(x), \phi(y)\rangle$ with $\phi$ being a two-layer MLP. \par 
    \textbf{Non-contrastive pre-training} Non-contrastive frameworks alleviate the need for negative sampling by directly maximizing the similarity between input graphs and their augmentation. We will utilize the simplest objective \cite{chen2020simple}
    \begin{align}\label{eqn: simsiam}
        \mathcal{L}_{\text{non-cst}} = \frac{1}{N} \sum_{G \in \mathbf{G}} \langle \psi(h(G)), \sg(\psi(h(G^+)) \rangle ,
    \end{align}
    where $\psi$ is a \emph{prediction head} that takes its form as a two-layer MLP and \sg\ stands for the stop-gradient operation. Both components have been shown necessary for avoiding collapsed solutions \cite{chen2021exploring}.\\
    The prevailing paradigm in graph data augmentations are of the \emph{perturbation} style, i.e., perturbing node attributes or perturbing the graph structure \cite{ding2022data}. It has been observed in previous works \cite{you2020graph} that the augmentation quality in self-supervised graph representation learning plays a vital role in downstream performance, most likely due to the different types of \emph{invariance} semantics encoded by the augmentation strategy.  In particular, the standard practice of graph structure perturbation techniques that adds or drops either nodes or edges essentially expresses a strong belief in invariance with respect to a small perturbation measured in terms of graph edit distance \cite{gao2010survey}. However, this may not provide reasonable invariance if the underlying graph is well-structured. For example, in the case of stochastic block model \cite{goldenberg2010survey}, randomly dropping or adding a significant proportion of edges changes the block structure and hence makes the precise semantics of the invariance questionable. Moreover, the inductive bias of the encoder also has an intriguing relationship between the augmentation strategy as well as downstream performances of self-supervised pre-training \cite{saunshi2022understanding}. Under the context of KGNNs, we intuitively expect that the learned hidden graphs shall somehow extract the \emph{signals} in the underlying graph instead of noise. Therefore, we require the graph data augmentation strategy to be \emph{structure-preserving}. To begin with, we will use the following random graph model \cite{goldenberg2010survey} as a starting point that describes the generating process of the underlying graph $G$, with each entry of its adjacency matrix sampled independently from a Bernoulli distribution,
    \begin{align}
        A_{uv} \iid \ber(\theta_{uv}), \forall u, v \in V(G).
    \end{align}
    The above formulation is nonparametric and very general, and covers many important generative mechanisms in random graph models, to name a few:
    \begin{description}
        \item[Graphon \cite{gao2015rate}] Suppose that each node $v \in V(G)$ is associated with a scalar $\xi_u \iid \unif(0, 1)$ that is independently drawn from a uniform distribution. And the generating probability is constructed via the \emph{graphon} $f: [0, 1]^2 \mapsto [0, 1]$, with $\theta_{uv} = f(\xi_u, \xi_v)$. 
        \item[Latent space model \cite{goldenberg2010survey}] We may alternatively let the unobserved node-wise characteristic be an arbitrary $d$-dimensional vector $\{\lambda_v\}_{v \in V(G)}$, and the generating probability is formulated as $\theta_{uv} = f(\lambda_u, \lambda_v)$, where with a slight abuse of notation we let $f$ be some continuous function that takes two $d$-dimensional vector arguments.
    \end{description}
    A notable result in modern random graph theory is that while precisely estimating the function $f$ in graphon or latent space models may encounter identifiability issues, the generating probabilities $\Theta = \{\theta_{uv}\}_{u, v \in V(G)}$ could be recovered in a provable fashion using the idea of \emph{universal singular value thresholding} (USVT) \cite{chatterjee2015matrix}. Inspired by this phenomenon, we use the estimated generating matrix $\widehat{\Theta}$ as a reference of \emph{graph structure}, and use this matrix to generate augmented views, which share (in an asymptotic fashion) similar graph structures with the input graph $G$. Since the process applies to many latent graph generative models, we term our augmentation method \textbf{L}atent \textbf{G}raph \textbf{A}ugmentation (LGA), detailed in algorithm \ref{alg: aug_svt}.
    \begin{algorithm}
        \caption{LGA using universal singular value thresholding (USVT)}
        \label{alg: aug_svt}
        \begin{algorithmic}[1]
            \Require Input graph $G = (V, E)$ with adjacency matrix $A$, threshold $\tau > 0$.
            \State Let $m$ be the number of nodes in $G$. Obtain the singular value decomposition of $A$: $A = \sum_{i}^m \sigma_i u_i v_i^T$
            \State Select indices based on the threshold $\mathcal{S} = \{i: \sigma_i \ge \tau \sqrt{m}\}$.
            \State Compute $\widetilde{\Theta} = \sum_{i \in \mathcal{S}} \sigma_i u_i v_i^T$
            \State Obtain the recovered random graph probability matrix $\widehat{\Theta}$ via clipping $\widetilde{\Theta}$:
            \begin{align}
                \widehat{\Theta}_{uv} = 
                \begin{cases}
                    \widetilde{\Theta}_{uv}\quad &\text{If $\widetilde{\Theta}_{uv} \in [0, 1]$ } \\
                    0 &\text{If $\widetilde{\Theta}_{uv} < 0$ } \\
                    1 &\text{If $\widetilde{\Theta}_{uv} > 1$ }
                \end{cases}
            \end{align}
            \State Generate matrix $A^+$ with each $A^+_{uv} \iid \ber(\widehat{\Theta}_{uv})$ with $u, v \in [m]$.\\
            \Return Positive sample $G^+$ derived from adjacency matrix $A^+$, i.e., with node features left untouched.
        \end{algorithmic}
    \end{algorithm}
    The most important step in LGA is to select a sufficiently large singular value based on a predefined (relative) threshold $\tau$, with its magnitude reflecting the trade-off between signal extraction and noise removal. In this paper, we cast the threshold $\tau$ as a tunable \emph{hyperparameter} which is regarded as a crude estimate of the upper bound of the matrix 
    $A - \Theta$ (normalized by the square root of graph size) that is shown to theoretically characterize the recovery under dense graph regimes \cite{chatterjee2015matrix}.
    \subsection{Complexity considerations}
    Now we present a brief analysis of the computational complexity of the proposed approach. Firstly, the complexity of obtaining the graph representation could be analyzed in a similar fashion as in \cite{nikolentzos2020random}: Ff we use $M$ hidden graphs with each involved in computing similarities up to order $P$ will incur a computational cost of $O(Md(nm + Pm^2 + Pn^2))$ for a dense sample graph and $O(Md(nm + Pm^2 + Ps))$ for a sparse sample graph with $s$ edges. The resulting complexity is comparable to a $P$-layer MPNN model with each layer producing hidden representations of dimension $m$. Next, we consider the cost of obtaining the characteristic matrix $B$ using graph diffusion. While we may use a truncated matrix sum with a moderate number of steps to reasonably approximate the diffusion matrix, we found in our experiments that typically a few diffusion steps (i.e., no more than three steps) suffice for performance improvements. Consequently, the resulting computational cost is dominated by that of representation computation. Finally, the computation cost of LGA (dominated by that of SVD factorization with $O(n^3)$) is only incurred once for each sample graph during training time. Under moderate-sized graphs, we argue that the computational cost is controllable.
    \section{Related works}
    \subsection{Neural models for graph representation learning}
    As graphs are generally believed to contain rich structural knowledge, neural architectures over graph-structured objects require different inductive biases compared to neural architectures over independently identically distributed (i.i.d.) data. 
    The authors in \cite{battaglia2018relational} proposed the concept of \emph{relational inductive bias} encoded by MPNNs that are closely related to dynamic programming. However, MPNNs are criticized for their limited expressivity \cite{morris2021weisfeiler, xu2018powerful}, we refer the readers to the article \cite{morris2021weisfeiler} for a thorough overview. 
    A more recent line of works derives neural architectures over graphs from the theory of graph kernels \cite{lei2017deriving, nikolentzos2020random, nikolentzos2023geometric}. In comparison to MPNNs, kernel approaches compute convolutions over graphs in a different way that might be regarded as a more transparent generalization of the convolution operation used in image modeling \cite{feng2022kergnns}. Besides, kernel approaches allow intuitive visualization of the learned hidden graphs (graph filters), thereby providing a natural sense of \emph{interpretability}.
    \subsection{Self-supervision on graphs}\label{rw: gcl}
    Early developments on self-supervised learning (SSL) on graphs considered using link prediction as the pretext task for downstream applications \cite{kipf2016variational}. 
    Later developments have been mostly focusing on applying SSL frameworks based on \emph{graph data augmentations (GDA)} \cite{velickovic2019deep, you2020graph, thakoor2021bootstrapped}
    , with the prevailing practices mostly involving editing graph structure (i.e., adding or dropping nodes or edges) and node features \cite{ding2022data} either randomly \cite{you2020graph} or non-randomly \cite{suresh2021adversarial, li2022let, lin2022spectral}.
    These modifications are based on the belief that the invariance relation is captured through certain edits of graphs. 
    \section{Experiments}
    In this section, we report empirical evaluations on benchmark datasets. The evaluations are conducted for both supervised learning tasks which focus on the empirical performance of the proposed SWAG network design, and self-supervised learning tasks which further investigate the effectiveness of the LGA mechanism. We also provide visualizations of the learned hidden graphs as an indirect way of comparing canonical patterns in the data, as well as an ablation study that inspects the effects of important hyperparameters.
    \subsection{Datasets}
    We use $8$ public datasets \cite{pinar2016deep}, with $4$ bio/chemo-informatics network datasets: MUTAG, D\&D, NCI1 and PROTEINS, and $4$ social interaction network datasets: IMDB-BINARY(IMDB-B), COLLAB, REDDIT-BINARY(RDT-B) and REDDIT-MULTI-5K(RDT-M5K). Across all the datasets, the evaluation task is graph classification with varying numbers of classes.
    For a more detailed introduction of the datasets as well their summary statistics, we refer to \cite{pinar2016deep} for details.
    \subsection{Baseline comparisons}
    As our experiments involve both supervised learning and self-supervised learning, we describe our baseline comparison strategy corresponding to both learning schemes as follows: \\
    \textbf{Supervised learning baselines} 
    We pick three representative graph kernel methods: shortest path kernel (SP) \cite{borgwardt2005shortest}, graphlet kernel \cite{shervashidze2009efficient} and Weisfeiler-leman subtree kernel \cite{shervashidze2011weisfeiler}. For MPNN methods, choose five influential MPNNs that applies to graph classification problems, namely DGCNN \cite{zhang2018end}, DiffPool \cite{ying2018hierarchical}, ECC \cite{simonovsky2017dynamic}, GIN \cite{xu2018powerful} and GraphSAGE \cite{hamilton2017inductive}. Additionally, we compare our proposed model with two recent KGNN models, RWGNN \cite{nikolentzos2020random} and GRWNN \cite{nikolentzos2023geometric}.\\
    \textbf{Self-supervised learning baselines} For a thorough investigation, we make two types of comparisons:
    \begin{description}
        \item[Comparison with MPNN encoders] We will compare our proposed \swag encoder with \lga augmentation scheme with various state-of-the-art self-supervised GRL frameworks: GraphCL \cite{you2020graph}, InfoGraph \cite{sun2019infograph}, GCC \cite{qiu2020gcc}, AD-GCL \cite{suresh2021adversarial}, RGCL \cite{li2022let}, GraphMAE \cite{hou2022graphmae} and GCL-SPAN \cite{lin2022spectral}, with their underlying encoder being a five-layer GIN \cite{xu2018powerful}. We note that such kind of comparisons are rarely done in previous works on KGNNs.
        \item[Comparison with KGNN encoders] Among the aforementioned baseline methods, four of them indeed apply to KGNN encoders. Including GraphCL \cite{you2020graph}, GCC \cite{qiu2020gcc}, RGCL \cite{li2022let} and GCL-SPAN \cite{lin2022spectral}. We additionally apply these augmentation strategies with a standard RWGNN encoder as a comparison of self-supervised KGNN models.
    \end{description}
    \begin{table*}[]
        \centering
        \caption{Experimental results on supervised graph classification  over $4$ molecular network datasets and $4$ social network datasets, reported with format \result{\text{\texttt{mean}}}{\text{\texttt{std}}}, with \texttt{mean} and \texttt{std} (abbreviation for standard deviation) computed under $10$ trials for each setting. We use {\color{red}\textbf{bolded red}} to highlight the best performance and {\color{blue}\underline{underlined blue}} to highlight the second best performance both in the sense of mean value.}
            \begin{tabular}{lcccccccc}
    \toprule
     & MUTAG & D\&D & NCI1 & PROTEINS & IMDB-B & COLLAB & RDT-B & RDT-M5K 
   \\
   \midrule
   SP & \result{80.2}{6.5} & \result{78.1}{4.1} & \result{72.7}{1.4} & \result{75.3}{3.8} & \result{57.7}{4.1} & \result{79.9}{2.7} & \result{89.0}{1.0} & \result{51.1}{2.2}
   \\
   GR & \result{80.8}{6.4} & \result{75.4}{3.4} & \result{61.8}{1.7} & \result{71.6}{3.1} & \result{63.3}{2.7} & \result{71.1}{1.4} & \result{76.6}{3.3} & \result{38.1}{2.3}
   \\
   WL & \result{84.6}{8.3} & \results{78.1}{2.4} & \resultf{84.8}{2.5} & \result{73.8}{4.4} & \result{72.8}{4.5} & \resultf{78.0}{2.0} & \result{74.9}{1.8} & \result{49.6}{2.0}
   \\
   \midrule
   DGCNN & \result{84.0}{6.7} & \result{76.6}{4.3} & \result{76.4}{1.7} & \result{72.9}{3.5} & \result{69.2}{3.0} & \result{71.2}{1.9} & \result{87.8}{2.5} & \result{49.2}{1.2}
   \\
   DiffPool & \result{79.8}{7.1} & \result{75.0}{3.5} & \result{76.9}{1.9} & \result{73.7}{3.5} & \result{68.4}{3.3} & \result{68.9}{2.0} & \result{89.1}{1.6} & \result{53.8}{1.4}
   \\
   GIN & \result{84.7}{6.7} & \result{75.3}{2.9} & \results{80.0}{1.4} & \result{73.3}{4.0} &\result{71.2}{3.9} & \results{75.6}{2.3} & \result{89.9}{1.9} & \results{56.1}{1.7}
   \\
   GraphSAGE & \result{83.6}{9.6} & \result{72.9}{2.0} & \result{76.0}{1.8} & \result{73.0}{4.5} & \result{68.8}{4.5} & \result{73.9}{1.7} & \result{84.3}{1.9} & \result{50.0}{1.3} \\
   \midrule
   RWGNN    & \result{89.2}{4.3} & \result{77.6}{4.7} & \result{73.9}{1.3} & \result{74.7}{3.3} & \result{70.6}{4.4} & \result{71.9}{2.5} & \results{90.4}{1.9} & \result{53.4}{1.6}
   \\
   GRWNN    & \result{83.4}{5.6} & \result{75.6}{4.6} & \result{67.7}{2.2} & \result{74.9}{3.5} & \result{72.8}{4.2} & \result{72.1}{1.9} & \result{90.0}{1.8} & \result{54,4}{1.7} \\
   \midrule
   SWAG(SL)     & \results{89.4}{6.0} & \result{77.8}{3.8} & \result{73.6}{1.8} & \results{76.3}{4.4} & \results{73.1}{3.3} & \result{74.1}{1.6} & \result{90.3}{1.8} & \result{55.4}{2.2} \\
   SWAG(SSL+FT) & \resultf{90.3}{6.2} & \resultf{79.5}{2.3} & \result{76.2}{2.7} & \resultf{77.5}{4.5} & \resultf{74.4}{3.3} & \result{74.6}{2.5} & \resultf{91.5}{1.7} & \resultf{56.2}{1.7} 
   \\
   \bottomrule
\end{tabular}
        \label{tab: sl_performance}
    \end{table*}
    \subsection{Experimental setup}
    \noindent\textbf{Evaluation protocol} We perform $10$-fold cross-validation to obtain testing performance across all the models, with the same splitting configuration as in \cite{errica2019fair}. Within each fold, we use $10\%$ of the data as a hold-out validation set that enables model selection. We use accuracy as the evaluation metric across all the datasets.\\
    \textbf{Training configurations} Across all the experimental trials, we use a \swag network architecture as the representation extractor $h$, and a two-layer MLP with hidden dimension $32$ as the downstream predictor. As not all of the chosen datasets contain node features, for those without node features we use node degree as the node feature. For the graph diffusion transform, we use the personalized PageRank (PPR) configuration $ \beta_j = \alpha (1 - \alpha)^j $ with teleport probability $\alpha = 0.15$ as suggested in \cite{gasteiger2019diffusion}. We train for $500$ epochs using the Adam optimizer \cite{kingma2014adam} with a learning rate $0.01$, under a batch size of $64$. Across all the experimental trials involving KGNN, we tune the number of hidden graphs $M$ in the range $\{8, 16\}$, with the number of nodes in each hidden graph tuned in the range $\{5, 10\}$. For the node features corresponding to hidden graphs, we tune the feature dimension in the range $\{32, 64\}$ for chemo-informatic network datasets, and $\{4, 8\}$ for the social network datasets. We choose the walk length, as well as the number of graph diffusion steps to be no greater than $3$ for computational efficiency. During self-supervision, we tune the thresholding parameter $\tau$ in the USVT scheme inside the range $[0.3, 4.2]$, with a detailed study on this parameter elaborated in section \ref{sec: ablation}. We adopt both contrastive pre-training objective \eqref{eqn: infonce} with negative samples chosen as the remaining ones in the same batch for any anchor graph, and the non-contrastive pretraining objectives \eqref{eqn: simsiam}. For both objectives, the additionally involved neural functions $\phi$ and $\psi$, as explained in section \ref{sec: ssl}, are parameterized by a two-layer MLP with $32$ hidden units. During downstream task adaptation, we tested both probing and fine-tuning and report the better-performing one. \\
    \textbf{Implementations} We mostly base our model implementation on PyTorch \cite{paszke2019pytorch}. For all the baseline methods, we adopt open-source implementations available in the corresponding papers and inherit their default training configurations.
    \subsection{Performance on supervised graph classification}
    \begin{table*}[]
        \centering
        \caption{Experimental results on self-supervised graph classification over $4$ molecular network datasets and $4$ social network datasets, reported with format \result{\text{\texttt{mean}}}{\text{\texttt{std}}}, with \texttt{mean} and \texttt{std} (abbreviation for standard deviation) computed under $10$ trials for each setting. We use {\color{red}\textbf{bolded red}} to highlight the best performance and {\color{blue}\underline{underlined blue}} to highlight the second best performance both in the sense of mean value.}
            \begin{tabular}{lcccccccc}
    \toprule
     & MUTAG & D\&D & NCI1 & PROTEINS & IMDB-B & COLLAB & RDT-B & RDT-M5K 
   \\
   \midrule
   \multicolumn{9}{c}{With MPNN encoder} \\
   \midrule
   GraphCL & \result{81.5}{7.0} & \result{70.4}{3.8} & \result{72.4}{3.0} & \result{69.3}{3.6} & \result{64.7}{5.1} & \result{74.4}{1.3} & \result{83.4}{3.0} & \result{49.9}{1.3} \\
   GCC & \result{83.6}{7.4} & \result{67.9}{2.3} & \result{68.0}{2.0} & \result{67.2}{2.4} & \result{74.4}{4.0} & \result{77.5}{1.7} & \result{85.5}{3.1} & \result{48.7}{1.2} \\
   RGCL & \results{89.9}{6.2} & \result{78.8}{3.2} & \result{78.4}{2.1} & \result{75.5}{3.8} & \result{73.2}{4.0} & \result{70.8}{1.7} & \result{89.5}{2.5} & \result{56.2}{2.1} \\
   GCL-SPAN & \result{85.7}{7.9} & \result{79.0}{3.0} & \result{73.8}{2.0} & \result{74.2}{4.4} & \result{71.5}{3.1} & \result{71.3}{2.0} & \result{71.9}{3.0} & \result{53.7}{1.7} \\
   \midrule
   InfoGraph & \result{87.7}{6.7} & \resultf{79.5}{2.1} & \resultf{80.3}{1.8} & \result{74.1}{3.6} & \result{71.4}{2.9} & \result{75.4}{2.1} & \result{91.1}{1.5} & \resultf{56.5}{1.2}\\
   AD-GCL & \result{88.1}{1.4} & \result{74.8}{0.7} & \result{69.4}{0.6} & \result{73.6}{0.6} & \result{71.2}{0.5} & \result{73.3}{0.5} & \result{84.9}{1.4} & \result{54.9}{0.7} \\
   GraphMAE & \result{85.6}{1.4} & \result{72.9}{1.6} & \results{79.3}{0.5} & \result{75.7}{0.3} & \resultf{75.4}{0.6} & \resultf{80.2}{0.2} & \result{84.5}{0.5} & \result{52.9}{0.3}
   \\
   \midrule
   \multicolumn{9}{c}{With RWGNN encoder} \\
   \midrule
   GraphCL & \result{86.2}{6.3} & \result{75.1}{3.0} & \result{67.7}{2.8} & \result{71.0}{4.5} & \result{67.7}{4.4} & \result{73.1}{1.8} & \result{82.1}{4.9} & \result{52.9}{2.1} \\
   GCC & \result{84.7}{6.7} & \result{74.9}{4.8} & \result{67.0}{1.9} & \result{72.1}{3.0} & \result{71.1}{4.7} & \results{78.3}{1.6} & \result{85.1}{2.8} & \result{49.0}{1.1} \\
   RGCL & \result{83.5}{8.3} & \result{74.8}{3.8} & \result{64.2}{2.1} & \result{72.7}{2.6} & \result{70.9}{3.9} & \result{68.2}{2.4} & \result{82.9}{2.7} & \result{49.5}{1.9} \\
   GCL-SPAN & \result{85.2}{8.5} & \result{76.2}{2.8} & \result{62.6}{2.7} & \result{75.2}{4.5} & \result{71.4}{2.7} & \result{68.4}{1.9} & \result{79.4}{1.9} & \result{49.0}{2.3} \\
   \midrule
   SWAG+LGA(infonce) & \result{89.1}{6.7} & \result{78.0}{2.4} & \result{75.6}{3.6} & \results{76.3}{4.3} & \result{74.0}{2.9} & \result{74.6}{2.5} & \results{91.5}{2.5} & \results{56.2}{1.7}
   \\
   SWAG+LGA(simsiam) & \resultf{90.3}{6.2} & \results{79.5}{2.3} & \result{76.2}{2.7} & \resultf{77.5}{4.5} & \results{74.4}{3.3} & \result{74.1}{2.0} & \resultf{91.5}{1.7} & \result{56.0}{2.1} 
   \\
   \bottomrule
\end{tabular}
        \label{tab: ssl_performance}
    \end{table*}
    We present experimental evaluations on supervised graph classification in table \ref{tab: sl_performance}. We have the following observations:
    \begin{itemize}[leftmargin=*]
        \item Our proposed \swag model is demonstrated to achieve comparable or superior performance to the two KGNN baselines, measured in \emph{mean} performance.
        On datasets with more dense graphs present such as IMDB-B and COLLAB, we found the improvements to be more evident, which might be attributed to the topological information cascading provided by the diffusion operation in the \swag mechanism.
        \item When compared against MPNN and graph kernel baselines, \swag achieves comparable performance with SOTA models on $6$ of the $8$ datasets, with the results in MUTAG, PROTEINS, and IMDB-B datasets relatively convincing. Another notable fact is that for certain datasets like NCI$1$ and COLLAB, using WL kernels leads to better results than neural GRL primitives, which is understandable since graph-level discrimination tasks are more sensitive to the expressivity limit of MPNNs, which was explicitly achieved using WL kernels.
        \item We additionally report the performance of \swag under self-supervision with \lga augmentation, the result shows that for all the $8$ datasets, self-supervision is beneficial to model performance.
    \end{itemize}
    \begin{figure}
        \centering
        \subcaptionbox{\label{fig: protines_viz_1}}{%
         \includegraphics[width=.21\linewidth]{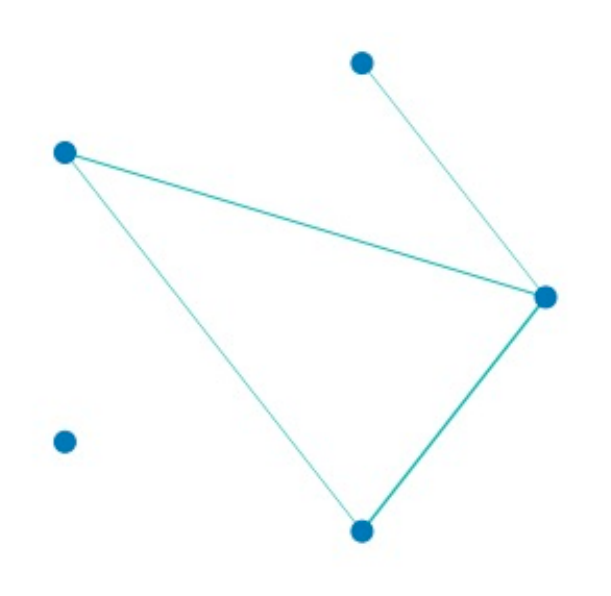}%
         }\hfill
        \subcaptionbox{\label{fig: protines_viz_2}}{%
         \includegraphics[width=.21\linewidth]{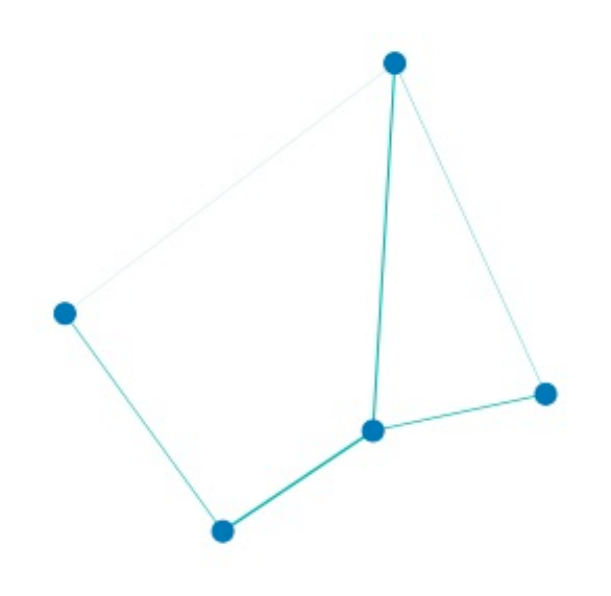}%
         }\hfill
         \subcaptionbox{\label{fig: protines_viz_3}}{%
         \includegraphics[width=.21\linewidth]{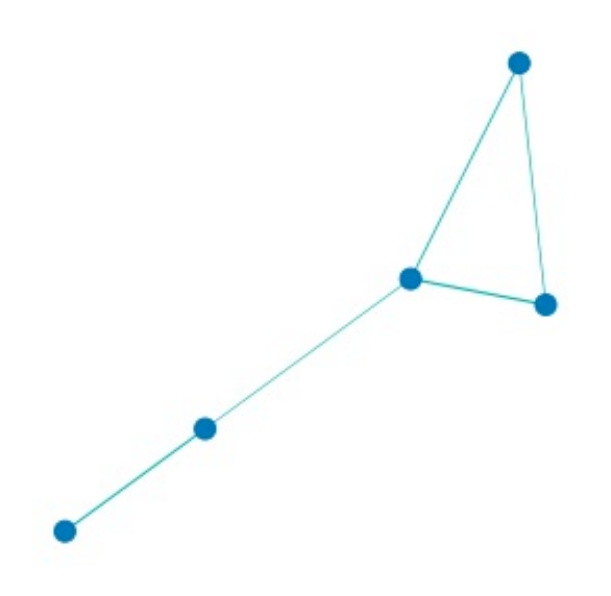}%
         }\hfill
         \subcaptionbox{\label{fig: protines_viz_4}}{%
         \includegraphics[width=.21\linewidth]{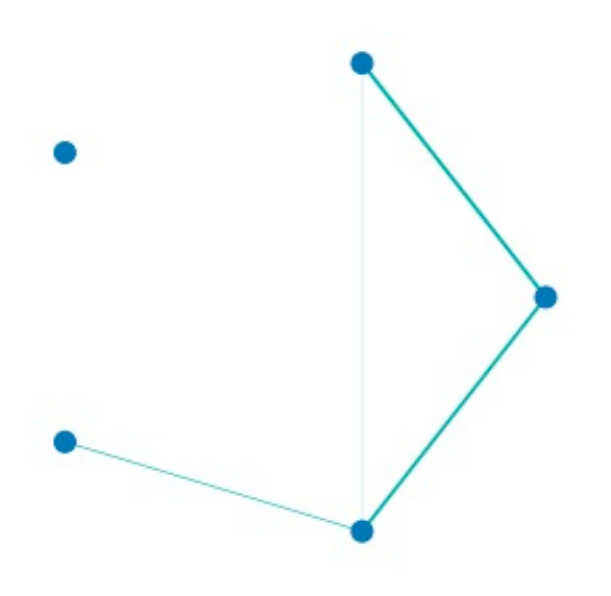}%
         }\hfill
        \subcaptionbox{\label{fig: rdt_viz_1}}{%
         \includegraphics[width=.21\linewidth]{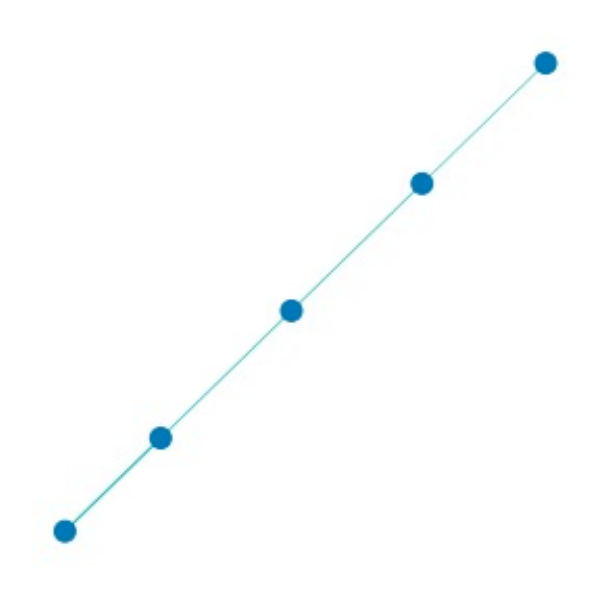}%
         }\hfill
         \subcaptionbox{\label{fig: rdt_viz_2}}{%
         \includegraphics[width=.21\linewidth]{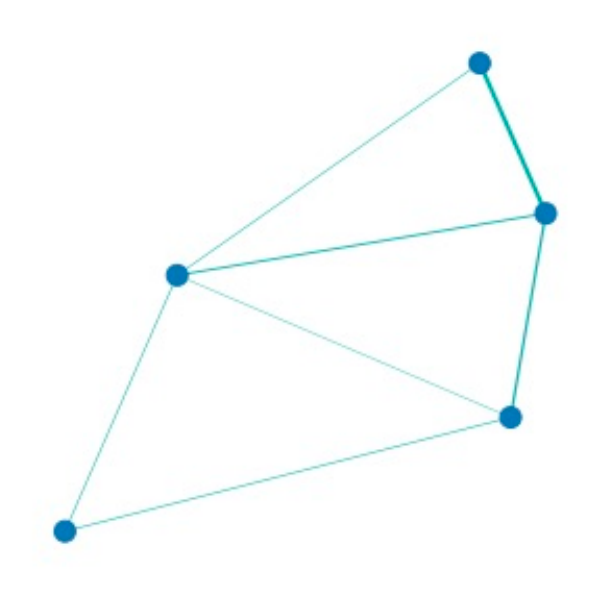}%
         }\hfill
         \subcaptionbox{\label{fig: rdt_viz_3}}{%
         \includegraphics[width=.21\linewidth]{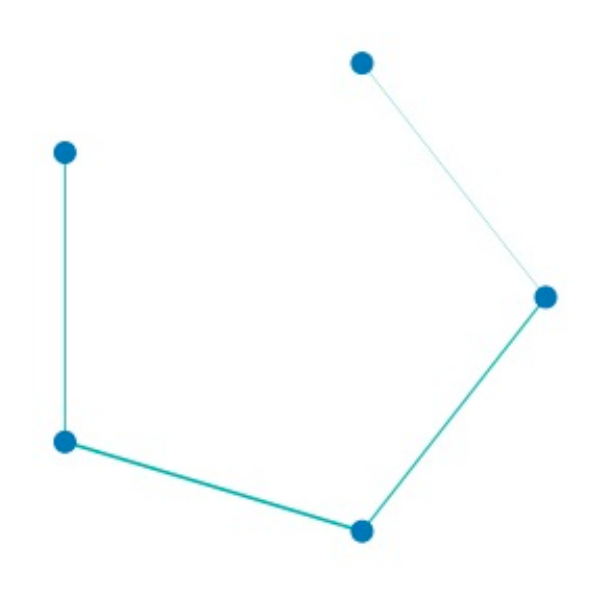}%
         }\hfill
         \subcaptionbox{\label{fig: rdt_viz_4}}{%
         \includegraphics[width=.21\linewidth]{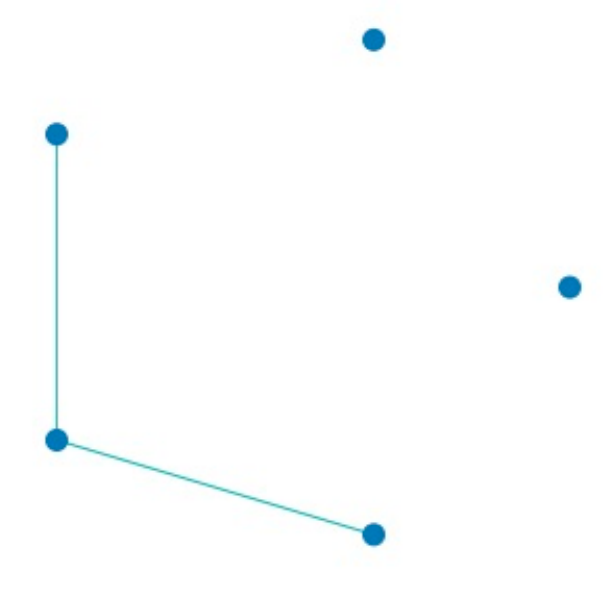}%
         }\hfill
        \caption{Visualization of $4$ randomly chosen learned subgraphs. Figure \ref{fig: protines_viz_1} through figure \ref{fig: protines_viz_4} are visualizations of hidden graphs learned from the PROTEINS dataset; Figure \ref{fig: rdt_viz_1} through figure \ref{fig: rdt_viz_4} are learned from the REDDIT-BINARY(RDT-B) dataset.}
        \label{fig: viz}
    \end{figure}
    \subsection{Performance on self-supervised graph classification}
    We present a detailed report comparing our approach and self-supervised GRL baselines in table \ref{tab: ssl_performance}, where we report separately the results under two different objectives. We summarize our findings as follows:
    \begin{itemize}[leftmargin=*]
        \item When compared against baselines with MPNN encoders, the proposed \swag model pre-trained with \lga augmentation achieves comparable results on $6$ of the $8$ datasets. 
        \item When compared against baselines with KGNN encoders, our proposed model exhibits a clear dominance over $7$ of the $8$ datasets, with the COLLAB dataset being the only exception where GCC is shown to be a better-performing augmentation method. We found this to be also consistent with comparisons against MPNN encoders where GCC is also quite effective. 
        \item We found two distinct self-supervision objectives under both contrastive and non-contrastive frameworks to behave similarly, therefore partially verifying that the performance gain of the proposed framework is mostly attributed to the \lga instead of self-supervision objectives.
    \end{itemize}

    \subsection{Visualizations of learned hidden graphs}\label{sec: viz}

    In this section, we assess the learned \swag model from the viewpoint of \emph{interpretability}. In particular, we choose two datasets, PROTEINS and RDT-B which are representative of bio/chemo-informatics networks as well as social networks. We randomly extract $4$ learned hidden graph from the learned \swag model, and plot them in figure \ref{fig: viz}. From the illustrations, it is visually clear that hidden graphs learned from distinct datasets exhibit different types of structural characteristics: Those learned from PROTEINS (the first row) show a relatively frequent pattern of \emph{rings} with typical instantiations being $3$-rings and $4$-rings. Meanwhile, those learned from RDT-B suggest an alternative common pattern that is somewhat \emph{chain-like}, i.e., figure \ref{fig: rdt_viz_1} and figure \ref{fig: rdt_viz_3} are both (isomorphic) $5$-chains. Therefore, the proposed \swag model offers reasonable transparency and interpretability that characterizes learned information, which is not possible for MPNNs.
    \subsection{Ablation study}\label{sec: ablation}
    \begin{figure}
        \centering
        \subcaptionbox{\label{fig: tau_proteins}}{%
         \includegraphics[width=.49\linewidth]{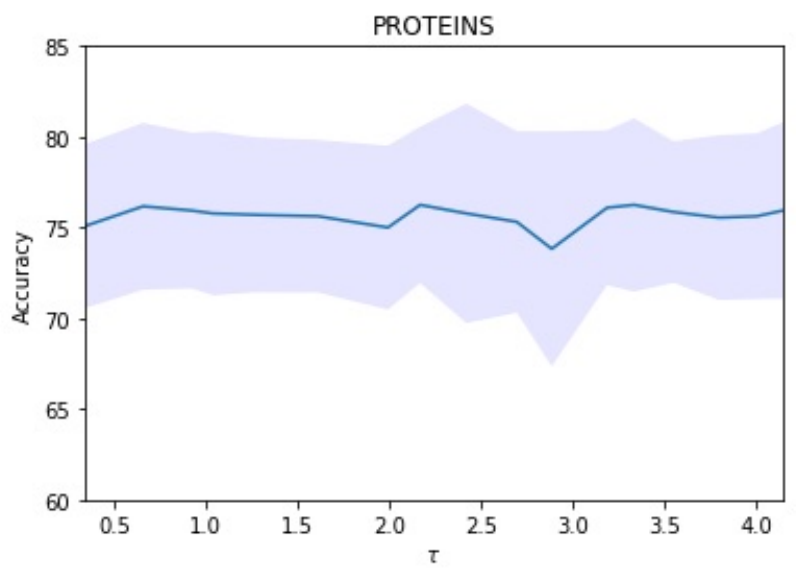}%
         }\hfill
        \subcaptionbox{\label{fig: tau_imdb}}{%
         \includegraphics[width=.49\linewidth]{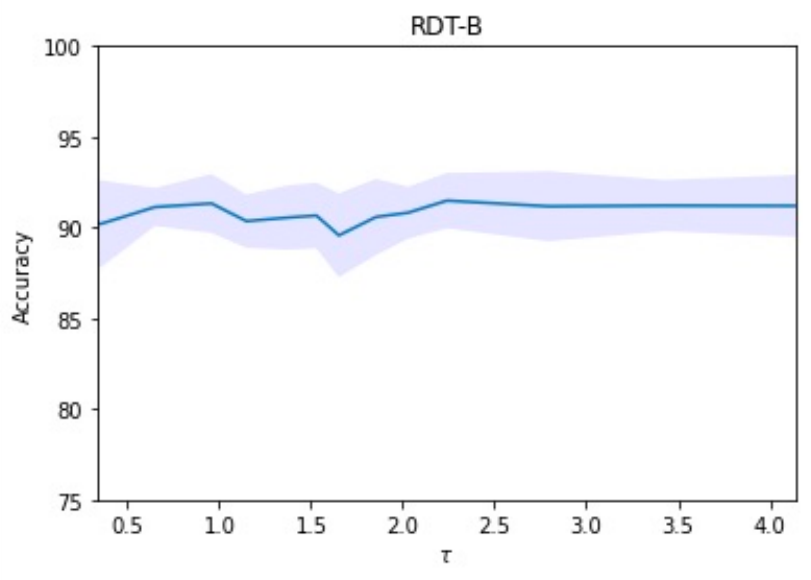}%
         }
        \caption{Investigation of model performance on PROTEINS (left) and RDT-B (right) when varying the USVT thresholding parameter $\tau$, mean performance are plotted along with shades indicating standard deviation under $10$-fold CV.}
        \label{fig: ablation_tau}
    \end{figure}
    In this section, we assess the effect of two critical hyperparameters in the proposed framework, namely the thresholding parameter $\tau$ involved in the USVT procedure, as well as the number of hidden graphs $M$ in the \swag architecture. All the experiments are conducted on two representative datasets, PROTEINS and RDT-B.\\
    \textbf{Effects of $\tau$} We draw insights from the renowned theoretical threshold $2.02$ which provably recovers dense random graphs as shown in \cite{chatterjee2015matrix}, and pick the largest $\tau$ value to be slightly bigger than twice the theoretical value as $4.2$. To accommodate for sparser graphs, we choose the smallest $\tau$ value to be slightly greater than $0$ for thorough evaluation, and report the results in figure \ref{fig: ablation_tau}. The results demonstrate only a mild level of fluctuation when varying $\tau$, suggesting that the structural invariance might be captured even with low-rank approximations of the input graph's adjacency matrix.\\
    \begin{figure}
        \centering
        \subcaptionbox{\label{fig: tau_proteins}}{%
         \includegraphics[width=.49\linewidth]{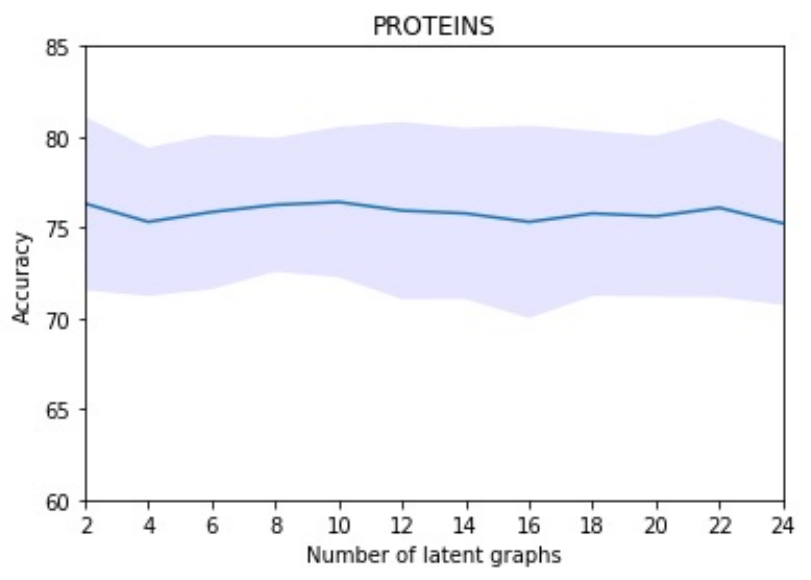}%
         }\hfill
        \subcaptionbox{\label{fig: tau_imdb}}{%
         \includegraphics[width=.49\linewidth]{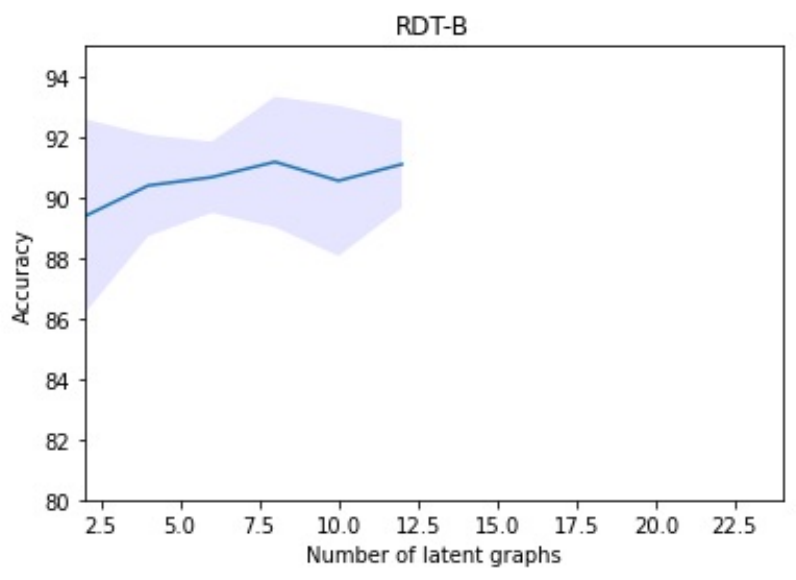}%
         }
        \caption{Investigation of model performance on PROTEINS (left) and RDT-B (right) when varying the number of hidden graphs $M$, mean performance are plotted along with shades indicating standard deviation under $10$-fold CV.}
        \label{fig: ablation_M}
    \end{figure}
    \textbf{Effects of $M$} As shown in section \ref{sec: viz}, the learned hidden graphs by the \swag architecture may exhibit recurring patterns (ignoring edge weights). Therefore, it is of interest to investigate whether a small number of hidden graphs suffice and whether too many hidden graphs hurts performance as they might end up being isomorphic structures. We tested the number of hidden graphs $M$ with a minimum of $2$ and a maximum of $24$ on the two datasets. The results are illustrated in figure \ref{fig: ablation_M}. The results suggest that while a relatively large number of hidden graphs (i.e., $M > 20$) may produce many similar learned graphs in terms of graph topology, the overall predictive performance is not significantly affected by varying $M$. 
    \section{Conclusion and future works}
    Several improvements are proposed regarding the design and learning of kernel graph neural networks (KGNNs). We extend previous formulations of random-walk graph neural networks into a more flexible framework and derive a novel neural architecture \swag that allows smoother optimization. We also initiate explorations on applying self-supervision to KGNNs, developing the structural-preserving graph data augmentation \lga which is beneficial to self-supervised KGNN learning. We believe this is just the initial step toward the development of better-performing KGNN models: It will be of great value if we may further bridge the theory and practice of MPNNs and KGNNs together, and create better supervised and self-supervised learning frameworks, we leave such promising research into future works.
    \bibliography{topology}

\begin{thebibliography}{10}

\bibitem{battaglia2018relational}
{\sc Battaglia, P.~W., Hamrick, J.~B., Bapst, V., Sanchez-Gonzalez, A., Zambaldi, V., Malinowski, M., Tacchetti, A., Raposo, D., Santoro, A., Faulkner, R., et~al.}
\newblock Relational inductive biases, deep learning, and graph networks.
\newblock {\em arXiv preprint arXiv:1806.01261\/} (2018).

\bibitem{borgwardt2005shortest}
{\sc Borgwardt, K.~M., and Kriegel, H.-P.}
\newblock Shortest-path kernels on graphs.
\newblock In {\em Fifth IEEE international conference on data mining (ICDM'05)\/} (2005), IEEE, pp.~8--pp.

\bibitem{chatterjee2015matrix}
{\sc Chatterjee, S.}
\newblock Matrix estimation by universal singular value thresholding.
\newblock {\em The Annals of Statistics\/} (2015), 177--214.

\bibitem{chen2020simple}
{\sc Chen, T., Kornblith, S., Norouzi, M., and Hinton, G.}
\newblock A simple framework for contrastive learning of visual representations.
\newblock In {\em International conference on machine learning\/} (2020), PMLR, pp.~1597--1607.

\bibitem{chen2021exploring}
{\sc Chen, X., and He, K.}
\newblock Exploring simple siamese representation learning.
\newblock In {\em Proceedings of the IEEE/CVF conference on computer vision and pattern recognition\/} (2021), pp.~15750--15758.

\bibitem{ding2022data}
{\sc Ding, K., Xu, Z., Tong, H., and Liu, H.}
\newblock Data augmentation for deep graph learning: A survey.
\newblock {\em ACM SIGKDD Explorations Newsletter 24}, 2 (2022), 61--77.

\bibitem{errica2019fair}
{\sc Errica, F., Podda, M., Bacciu, D., and Micheli, A.}
\newblock A fair comparison of graph neural networks for graph classification.
\newblock {\em arXiv preprint arXiv:1912.09893\/} (2019).

\bibitem{feng2022kergnns}
{\sc Feng, A., You, C., Wang, S., and Tassiulas, L.}
\newblock Kergnns: Interpretable graph neural networks with graph kernels.
\newblock In {\em Proceedings of the AAAI Conference on Artificial Intelligence\/} (2022), vol.~36, pp.~6614--6622.

\bibitem{gao2015rate}
{\sc Gao, C., Lu, Y., and Zhou, H.~H.}
\newblock Rate-optimal graphon estimation.
\newblock {\em The Annals of Statistics\/} (2015), 2624--2652.

\bibitem{gao2010survey}
{\sc Gao, X., Xiao, B., Tao, D., and Li, X.}
\newblock A survey of graph edit distance.
\newblock {\em Pattern Analysis and applications 13\/} (2010), 113--129.

\bibitem{gasteiger2019diffusion}
{\sc Gasteiger, J., Wei{\ss}enberger, S., and G{\"u}nnemann, S.}
\newblock Diffusion improves graph learning.
\newblock {\em Advances in neural information processing systems 32\/} (2019).

\bibitem{pmlr-v70-gilmer17a}
{\sc Gilmer, J., Schoenholz, S.~S., Riley, P.~F., Vinyals, O., and Dahl, G.~E.}
\newblock Neural message passing for quantum chemistry.
\newblock In {\em Proceedings of the 34th International Conference on Machine Learning\/} (International Convention Centre, Sydney, Australia, 06--11 Aug 2017), D.~Precup and Y.~W. Teh, Eds., vol.~70 of {\em Proceedings of Machine Learning Research}, PMLR, pp.~1263--1272.

\bibitem{gligorijevic2021structure}
{\sc Gligorijevi{\'c}, V., Renfrew, P.~D., Kosciolek, T., Leman, J.~K., Berenberg, D., Vatanen, T., Chandler, C., Taylor, B.~C., Fisk, I.~M., Vlamakis, H., et~al.}
\newblock Structure-based protein function prediction using graph convolutional networks.
\newblock {\em Nature communications 12}, 1 (2021), 3168.

\bibitem{goldenberg2010survey}
{\sc Goldenberg, A., Zheng, A.~X., Fienberg, S.~E., Airoldi, E.~M., et~al.}
\newblock A survey of statistical network models.
\newblock {\em Foundations and Trends{\textregistered} in Machine Learning 2}, 2 (2010), 129--233.

\bibitem{hamilton2017inductive}
{\sc Hamilton, W., Ying, Z., and Leskovec, J.}
\newblock Inductive representation learning on large graphs.
\newblock {\em Advances in neural information processing systems 30\/} (2017).

\bibitem{hou2022graphmae}
{\sc Hou, Z., Liu, X., Cen, Y., Dong, Y., Yang, H., Wang, C., and Tang, J.}
\newblock Graphmae: Self-supervised masked graph autoencoders.
\newblock In {\em Proceedings of the 28th ACM SIGKDD Conference on Knowledge Discovery and Data Mining\/} (2022), pp.~594--604.

\bibitem{kingma2014adam}
{\sc Kingma, D.~P., and Ba, J.}
\newblock Adam: A method for stochastic optimization.
\newblock {\em arXiv preprint arXiv:1412.6980\/} (2014).

\bibitem{kipf2016semi}
{\sc Kipf, T.~N., and Welling, M.}
\newblock Semi-supervised classification with graph convolutional networks.
\newblock {\em arXiv preprint arXiv:1609.02907\/} (2016).

\bibitem{kipf2016variational}
{\sc Kipf, T.~N., and Welling, M.}
\newblock Variational graph auto-encoders.
\newblock {\em CoRR abs/1611.07308\/} (2016).

\bibitem{lei2017deriving}
{\sc Lei, T., Jin, W., Barzilay, R., and Jaakkola, T.}
\newblock Deriving neural architectures from sequence and graph kernels.
\newblock In {\em International Conference on Machine Learning\/} (2017), PMLR, pp.~2024--2033.

\bibitem{li2022let}
{\sc Li, S., Wang, X., Zhang, A., Wu, Y., He, X., and Chua, T.-S.}
\newblock Let invariant rationale discovery inspire graph contrastive learning.
\newblock In {\em International Conference on Machine Learning\/} (2022), PMLR, pp.~13052--13065.

\bibitem{lin2022spectral}
{\sc Lin, L., Chen, J., and Wang, H.}
\newblock Spectral augmentation for self-supervised learning on graphs.
\newblock {\em arXiv preprint arXiv:2210.00643\/} (2022).

\bibitem{morris2021weisfeiler}
{\sc Morris, C., Lipman, Y., Maron, H., Rieck, B., Kriege, N.~M., Grohe, M., Fey, M., and Borgwardt, K.}
\newblock Weisfeiler and leman go machine learning: The story so far.
\newblock {\em arXiv preprint arXiv:2112.09992\/} (2021).

\bibitem{nikolentzos2020random}
{\sc Nikolentzos, G., and Vazirgiannis, M.}
\newblock Random walk graph neural networks.
\newblock {\em Advances in Neural Information Processing Systems 33\/} (2020), 16211--16222.

\bibitem{nikolentzos2023geometric}
{\sc Nikolentzos, G., and Vazirgiannis, M.}
\newblock Geometric random walk graph neural networks via implicit layers.
\newblock In {\em International Conference on Artificial Intelligence and Statistics\/} (2023), PMLR, pp.~2035--2053.

\bibitem{oord2018representation}
{\sc Oord, A. v.~d., Li, Y., and Vinyals, O.}
\newblock Representation learning with contrastive predictive coding.
\newblock {\em arXiv preprint arXiv:1807.03748\/} (2018).

\bibitem{paszke2019pytorch}
{\sc Paszke, A., Gross, S., Massa, F., Lerer, A., Bradbury, J., Chanan, G., Killeen, T., Lin, Z., Gimelshein, N., Antiga, L., et~al.}
\newblock Pytorch: An imperative style, high-performance deep learning library.
\newblock {\em Advances in neural information processing systems 32\/} (2019).

\bibitem{qiu2020gcc}
{\sc Qiu, J., Chen, Q., Dong, Y., Zhang, J., Yang, H., Ding, M., Wang, K., and Tang, J.}
\newblock Gcc: Graph contrastive coding for graph neural network pre-training.
\newblock In {\em Proceedings of the 26th ACM SIGKDD international conference on knowledge discovery \& data mining\/} (2020), pp.~1150--1160.

\bibitem{saunshi2022understanding}
{\sc Saunshi, N., Ash, J., Goel, S., Misra, D., Zhang, C., Arora, S., Kakade, S., and Krishnamurthy, A.}
\newblock Understanding contrastive learning requires incorporating inductive biases.
\newblock In {\em International Conference on Machine Learning\/} (2022), PMLR, pp.~19250--19286.

\bibitem{vishwanathan2010graph}
{\sc Schraudolph, N.~N., Kondor, R., and Borgwardt, K.~M.}
\newblock Graph kernels.
\newblock {\em Journal of Machine Learning Research 11\/} (2010), 1201--1242.

\bibitem{shervashidze2011weisfeiler}
{\sc Shervashidze, N., Schweitzer, P., Van~Leeuwen, E.~J., Mehlhorn, K., and Borgwardt, K.~M.}
\newblock Weisfeiler-lehman graph kernels.
\newblock {\em Journal of Machine Learning Research 12}, 9 (2011).

\bibitem{shervashidze2009efficient}
{\sc Shervashidze, N., Vishwanathan, S., Petri, T., Mehlhorn, K., and Borgwardt, K.}
\newblock Efficient graphlet kernels for large graph comparison.
\newblock In {\em Artificial intelligence and statistics\/} (2009), PMLR, pp.~488--495.

\bibitem{simonovsky2017dynamic}
{\sc Simonovsky, M., and Komodakis, N.}
\newblock Dynamic edge-conditioned filters in convolutional neural networks on graphs.
\newblock In {\em Proceedings of the IEEE conference on computer vision and pattern recognition\/} (2017), pp.~3693--3702.

\bibitem{stokes2020deep}
{\sc Stokes, J.~M., Yang, K., Swanson, K., Jin, W., Cubillos-Ruiz, A., Donghia, N.~M., MacNair, C.~R., French, S., Carfrae, L.~A., Bloom-Ackermann, Z., et~al.}
\newblock A deep learning approach to antibiotic discovery.
\newblock {\em Cell 180}, 4 (2020), 688--702.

\bibitem{sun2019infograph}
{\sc Sun, F.-Y., Hoffmann, J., Verma, V., and Tang, J.}
\newblock Infograph: Unsupervised and semi-supervised graph-level representation learning via mutual information maximization.
\newblock {\em arXiv preprint arXiv:1908.01000\/} (2019).

\bibitem{suresh2021adversarial}
{\sc Suresh, S., Li, P., Hao, C., and Neville, J.}
\newblock Adversarial graph augmentation to improve graph contrastive learning.
\newblock {\em Advances in Neural Information Processing Systems 34\/} (2021), 15920--15933.

\bibitem{thakoor2021bootstrapped}
{\sc Thakoor, S., Tallec, C., Azar, M.~G., Munos, R., Veli{\v{c}}kovi{\'c}, P., and Valko, M.}
\newblock Bootstrapped representation learning on graphs.
\newblock In {\em ICLR 2021 Workshop on Geometrical and Topological Representation Learning\/} (2021).

\bibitem{velickovic2019deep}
{\sc Velickovic, P., Fedus, W., Hamilton, W.~L., Li{\`o}, P., Bengio, Y., and Hjelm, R.~D.}
\newblock Deep graph infomax.
\newblock {\em ICLR (Poster) 2}, 3 (2019), 4.

\bibitem{wu2021self}
{\sc Wu, L., Lin, H., Tan, C., Gao, Z., and Li, S.~Z.}
\newblock Self-supervised learning on graphs: Contrastive, generative, or predictive.
\newblock {\em IEEE Transactions on Knowledge and Data Engineering\/} (2021).

\bibitem{xu2018powerful}
{\sc Xu, K., Hu, W., Leskovec, J., and Jegelka, S.}
\newblock How powerful are graph neural networks?
\newblock {\em arXiv preprint arXiv:1810.00826\/} (2018).

\bibitem{pinar2016deep}
{\sc Yanardag, P., and Vishwanathan, S.}
\newblock Deep graph kernels.
\newblock In {\em Proceedings of the 21th ACM SIGKDD International Conference on Knowledge Discovery and Data Mining\/} (New York, NY, USA, 2015), KDD '15, Association for Computing Machinery, p.~1365–1374.

\bibitem{ying2018hierarchical}
{\sc Ying, Z., You, J., Morris, C., Ren, X., Hamilton, W., and Leskovec, J.}
\newblock Hierarchical graph representation learning with differentiable pooling.
\newblock {\em Advances in neural information processing systems 31\/} (2018).

\bibitem{you2020graph}
{\sc You, Y., Chen, T., Sui, Y., Chen, T., Wang, Z., and Shen, Y.}
\newblock Graph contrastive learning with augmentations.
\newblock {\em Advances in neural information processing systems 33\/} (2020), 5812--5823.

\bibitem{zhang2018end}
{\sc Zhang, M., Cui, Z., Neumann, M., and Chen, Y.}
\newblock An end-to-end deep learning architecture for graph classification.
\newblock In {\em Proceedings of the AAAI conference on artificial intelligence\/} (2018), vol.~32.

\end{thebibliography}
    \bibliographystyle{acm}
\end{document}